\pdfoutput=1

\documentclass[11pt]{article}

\usepackage[]{acl}

\usepackage{times}
\usepackage{latexsym}
\usepackage{graphicx}
\usepackage{makecell}
\usepackage{multirow}
\usepackage{amsmath}
\usepackage{amsfonts}
\usepackage[T1]{fontenc}

\usepackage[utf8]{inputenc}

\usepackage{microtype}

\title{Flat Multi-modal Interaction Transformer for Named Entity Recognition}

\author{\hspace{2em} Junyu Lu$^\clubsuit$$^\spadesuit$ \hspace{2em} Dixiang Zhang$^\clubsuit$ \hspace{2em} Jiaxing Zhang$^\spadesuit$ \hspace{2em}  Pingjian Zhang$^\clubsuit$\thanks{\texttt{\ 
 Corresponding author}}\\
 $^\clubsuit$South China University of Technology \\
 $^\spadesuit$International Digital Economy Academy (IDEA)\\
 \texttt{lujunyu@idea.edu.cn} \hspace{2em}\\
 \texttt{sedxzhang@mail.scut.edu.cn} \\
 \texttt{pjzhang@scut.edu.cn}
 \\}

\begin{document}
\maketitle
\begin{abstract}
Multi-modal named entity recognition (MNER) aims at identifying entity spans and recognizing their categories in social media posts with the aid of images. However, in dominant MNER approaches, the interaction of different modalities is usually carried out through the alternation of self-attention and cross-attention or over-reliance on the gating machine, which results in imprecise and biased correspondence between fine-grained semantic units of text and image. To address this issue, we propose a Flat Multi-modal Interaction Transformer (FMIT) for MNER. Specifically, we first utilize noun phrases in sentences and general domain words to obtain visual cues. Then, we transform the fine-grained semantic representation of the vision and text into a unified lattice structure and design a novel relative position encoding to match different modalities in Transformer. Meanwhile, we propose to leverage entity boundary detection as an auxiliary task to alleviate visual bias. Experiments show that our methods achieve the new state-of-the-art performance on two benchmark datasets.
\end{abstract}

\section{Introduction}
Named entity recognition (NER) is a fundamental task in the field of information extraction, which involves determining entity boundaries from free text and classifying them into pre-defined categories, such as person (PER), location (LOC), and organization (ORG)~\citep{zhao2021modeling}. Along with the rapid development of social media, multi-modal deep learning is widely applied in the structured extraction from massive multimedia news and web product information~\citep{zhang2020multi,ju2020transformer}. As an important research branch of NER, multi-modal named entity recognition (MNER) significantly extends the text-based NER by taking the \{Sentence, Image\} pair as inputs~\citep{lu2018visual,kruengkrai2020improving, dosovitskiy2020image, 9747303}. Since the visual context associated with text content has been confirmed to help resolve the recognition of ambiguous multi-sense words and out-of-vocabulary words, MNER plays an important role in extracting entities from user-generated content on social media platforms such as Twitter ~\citep{li2015tweet}.

\begin{figure}[t]

\begin{center}
\centerline{\includegraphics[width=0.85 \linewidth]{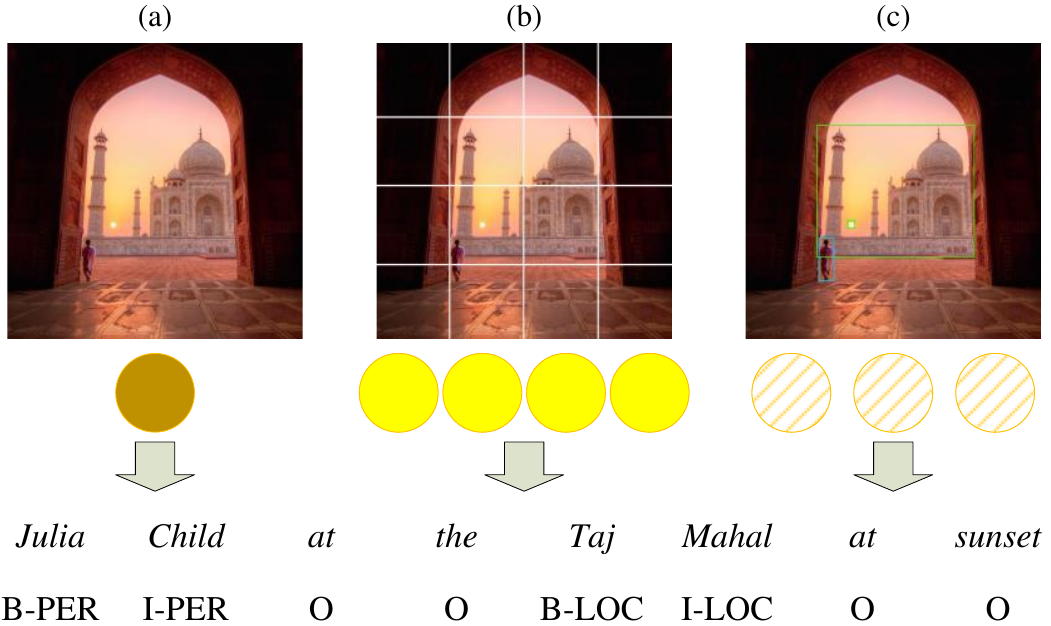}}
\vskip -0.1in
\caption{An example for multi-modal named entity recognition with different visual cues: (a) the whole image, (b) averagely segmented visual feature and (c) targeted visual objects.}\label{fig:systemodel}
\end{center}
\vskip -0.3in
\end{figure}

It has been the core issue in MNER to fully exploit the effective visual information and suppress the interference information, which directly affects the model performance. To this end, there are three lines of methods to integrate visual information into NER. (1) The first line is to encode the whole image into a global feature vector (Figure \ref{fig:systemodel}(a)) for augmenting each word representation~\citep{moon2018multimodal}. (2) The second line is to divide the feature map extracted from the whole image into multiple regions averagely (Figure \ref{fig:systemodel}(b)), which is the most dominant method currently. The method guides the word to learn a vision-aware representation through co-attention and gating mechanism~\citep{lu2018visual,zhang2018adaptive}, or uses a Transformer framework based on the combination of self-modal and cross-modal attention to interact textual and visual information~\citep{yu2020improving, sun2021rpbert}. (3) The last line is to use noun phrases to detect the image bounding boxes (Figure \ref{fig:systemodel}(c)) and fuse fine-grained words and visual objects by graph neural networks (GNN) ~\citep{zhang2021multi}.

Despite the success, the above works may not precisely exploit the fine-grained semantic correspondence between semantic units in an input sentence-image pair. Specifically, as shown in Figure \ref{fig:systemodel}, we believe that the crux of the issue lies in two aspects. (1) The global clue provided by image (a) and regional feature maps provided by images (b) are both implicit and vague, which are difficult to fit into fine-grained words. Previous practice tends to map the visual and textual representations into different spaces and then fuse them adaptively. However, the indirectness of information interaction through cross-modal attention and gating mechanism will lead to asymmetry in information acquisition~\citep{sun2021rpbert, yu2020improving}. (2) Specific visual objects derived from noun phrases are overly targeted, which makes some non-entities embodied in the images identified as entities incorrectly. Generally speaking, this kind of explicit information is helpful to identify some words as the correct entity type, such as ``\textit{the Taj Mahal}''. However, the visual objects prominent in the image may easily misidentified as entities, such as “\textit{sunset}”.

To handle the aforementioned issues, we propose a novel Flat Multi-model Interaction Transformer for MNER. The key insight comes from the lattice structure in Chinese NER~\citep{zhang2018chinese,li2020flat}, where word sequence is used as additional information to enhance the character representation. To fully exploit the available visual information, we use a combination of two visual objects, extracted from the whole image and derived from the noun phrases. We first represent the input sentence and image with a unified flat lattice structure consisting of fine-grained semantic units. Each unit corresponds to a word or visual object and its position. Meanwhile, inspired by the strategies of position representation, we design an ingenious position encoding for our flat lattice structure~\citep{shaw2018self,dai2019transformer}. In detail, we assign two positional indices for a unit: head position and tail position, by which we can correspond the visual object to the associated words span. Based on the flat lattice structure, we then resort to the fully-connected self-attention structure and long-distance dependencies modeling capability in Transformer~\citep{vaswani2017attention} to build bridges in the interaction between self-modal and cross-modal units. Finally, we utilize the CRF decoder to obtain the predicted labels. To largely eliminate the bias of visual context, we further introduce the entity boundary detection (EBD) as an auxiliary task.

We conduct extensive experiments on Twitter 2015 and Twitter 2017 benchmark datasets. The state-of-the-art performance and efficiency demonstrate the effectiveness of our methods.

\section{Related Work}

\paragraph{Multi-modal NER.} As an important role in many downstream NLP tasks,
including information retrieval ~\citep{chen2015event}, relation extraction ~\citep{miwa2016end} and question answering system ~\citep{diefenbach2018core}, the text-based NER task has attracted much attention in the journalistic and social fields. Deep learning approaches such as CNN, LSTM, attention mechanism and pre-trained models have achieved significant success in NER, by which we can effectively uncover and combine the character, word and sentence information in text sequence~\citep{ma2016end,akbik2019pooled,luo2020hierarchical}. Influenced by the extensive applications of multi-modal learning in neural machine translation, visual question answering and emotion detection~\citep{zhang2020multi,yin2020novel,gao2019dynamic}, many researches have focused on constructing multi-modal NER datasets and exploring the methods to guide entity recognition using images. The main idea of these early methods is encoding the text through LSTM and the image through pre-trained CNN, then implicitly interacting the information of two modalities~\citep{lu2018visual,moon2018multimodal,zhang2018adaptive}. Recently, ~\citep{yu2020improving} leverage BERT to model text sequence and creatively design a multi-modal interaction module based on Transformer and gating mechanism to perform self-modal and cross-modal information interaction alternately. ~\citep{zhang2021multi} further represent the input sentence-image pair as a unified graph to capture the various semantic relationships and introduce an extended GNN to conduct graph encoding via multi-modal semantic interactions. Different from above studies, our approach aims at representing the fine-grained semantic units of the text and image as a unified flat lattice structure. We further design a novel relative position encoding strategy to directly capture the interaction between different modalities in Transformer.

\paragraph{Lattice Structure.} Since word information is potentially useful for character-based Chinese NER task, the lattice structure is proposed for injecting the word information into the associated characters. Specifically, ~\citep{zhang2018chinese} first proposes Lattice-LSTM to explicitly exploit word boundary information, in which the matched lexical words are integrated into characters via a directed acyclic graph. Later, to overcome the limitations of the lattice structure so that it can be flexibly exploited, ~\citep{sui2019leverage} converts the lattice into graph and designs a collaborative graph network for encoding. ~\citep{zhao2021dynamic} proposes a dynamic cross- and self-lattice attention network to model dense interactions over word-character pairs.

\section{Method}

\begin{figure*}[t]
\begin{center}
\centerline{\includegraphics[width=0.95 \linewidth]{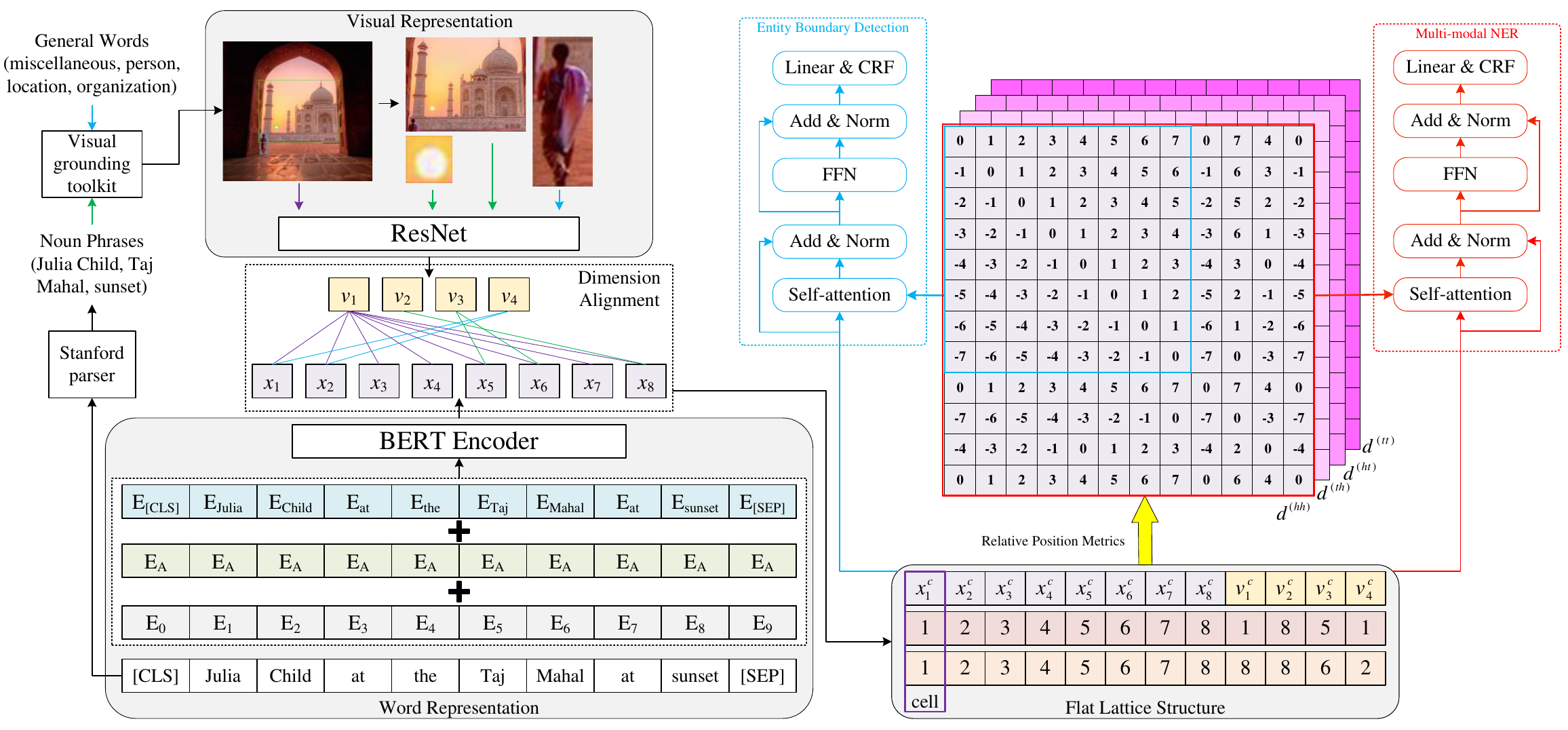}}
\vskip -0.1in
\caption{The overall architecture of our Flat Multi-modal Interaction Transformer (FMIT). On the right part, the blue-frame demonstrates the auxiliary task of entity boundary detection, and the red-frame demonstrates the main task of MNER.}\label{fig:FMIT}
\end{center}
\vskip -0.3in
\end{figure*}

Figure \ref{fig:FMIT} illustrates the overall architecture of our FMIT, which contains three main components: (1) Unified flat lattice structure for representing the input sentence-image pairs. (2) Transformer Encoder with relative position encoding method for interacting multi-modal information. (3) Training with entity boundary detection as an auxiliary task.

\paragraph{Task Formulation.} Given a sentence $S$  and its associated image $O$ as input, the goal of MNER is to extract a set of entities from $S$ and classify each extracted entity into one of the pre-defined categories. As with most existing work in MNER, we formulate the task as a sequence labelling problem. Let $S=(w_1,w_2,...,w_n)$ denote a sequence of input words, where $w_i$ with $i=1,2,...n$ denotes the $i$th word in the sentence and n represents the length of the sentence, and $Y=(y_1,y_2,...,y_n)$ be the corresponding entity labels for all words, where $y_1\in\mathcal Y$ and $\mathcal Y$ is the pre-defined label set with standard BIO schema ~\citep{sang1999representing}. We also use $O=(o_1,o_2,...,o_m)$ to denote a set of input visual objects of number $m$.

\subsection{Unified Flat Lattice Structure}
In this section, we take the sentence and image shown in Figure \ref{fig:systemodel} as an example to describe how to extract features from them and represent them with a flat lattice structure.

\paragraph{Word Representations.} Due to the capability of providing different representations for the same word in different contexts, we utilize pre-trained language model BERT ~\citep{devlin2019bert} as our sentence encoder. Following BERT, the input sentence is preprocessed by inserting the special token [CLS] and [SEP] at the beginning and the end positions, respectively. $S$ is then fed to BERT encoder to obtain the vectorized representation $H_x=(x_1,x_2,...x_n)$, where $x_i\in\mathbb R^{d_w}$ is the generated contextualized vector for $w_i$.

\paragraph{Visual Representations.} To capture the visual objects in $O$, except for employing the whole image, we also need to derive additional visual objects from the text. Similar to ~\citep{yin2020novel}, we use the constituency parsing tool in the Stanford parser to identity all noun phrases in the input sentence, and then apply a visual grounding toolkit ~\citep{yang2019fast} to detect bounding boxes (visual objects) for each noun phrase. Since it is difficult to use noun phrases merely to completely detect all potential visual objects in the image, according to the property of NER, we further introduce four general words of the pre-defined categories (i.e., miscellaneous, person, location and organization) to discover more relevant objects.

    To extract meaningful feature representations from images, we leverage a pre-trained 152-layer ResNet ~\citep{he2016deep} as a feature detector. We feed each visual object in $O$ to the ResNet and take the last hidden layer output as vectorized representation $H_v=(v_1,v_2,...,v_m)$, where $v_i\in\mathbb R^{d_v}$ is the generated visual representation for $o_i$.

\paragraph{Flat Lattice Construction.} The flat lattice structure aims to integrate the intra-modal and inter-modal information in a unified space, and enhances information coupling through a unique positioning scheme. Before representing the words and visual objects in a uniform lattice cell, we introduce two non-linear transformations with ReLU activation function to project different representations onto the same dimension:

\vskip -0.2in
{\small
\begin{align}
    x_i^c &=W_0(\operatorname{ReLU}(W_1 x_i+b_1))+b_0 \\
    v_i^c &=W_0(\operatorname{ReLU}(W_2 v_i+b_2 ))+b_0,
\end{align}
}
\vskip -0.1in
\noindent where $W_1\in\mathbb R^{d\times d_w}$, $W_2\in\mathbb R^{d\times d_v}$, $W_0\in\mathbb R^{d\times d}$ are weight matrices, and $b_1$, $b_2$, $b_0$ are scaler bias. $d$ is the dimension of unified representations of two modalities in the flat lattice.

To convert two independent sets of modalities to a flat lattice structure, we concatenate the aligned word representation and visual representation, and then flatten them into a unified sequence $E=([CLS],x_1^c,...,x_n^c,[SEP],v_1^c,...,v_m^c)$. As shown in Figure \ref{fig:FMIT}, the flat lattice can be defined as a set of cells, and a cell corresponds to a fine-grained semantic unit, a head and a tail. Specifically, for the word, its head and tail are equal, both indexes of the absolute position in $E$. For the visual object, when it is derived from a noun phrase, its head and tail are indexes of the first and last words of the corresponding noun phrase; in particular, when it is the whole image or derived from general words, we denote its head and tail as indexes of the first and last words of the sequence.
\vskip -0.6in

\subsection{Flat Multi-modal Interaction Transformer}

\paragraph{Relative Position Encoding for Flat Lattice Structure.} The flat lattice structure consists of cells with different modalities and different position intervals in the visual modality. As illustrated on the red-frame of Figure \ref{fig:FMIT}, to leverage the Transformer framework to encode interactions among cells, we design a relative position encoding for the cells. Specifically, for two cells $c_i$ and $c_j$ in the lattice, we consider two kinds of relations between them: intra-modal and inter-modal. Instead of directly encoding input with absolute position as vanilla Transformer, we calculate a dense vector to represent relative position by continuous transformation of the head and tail information. In this way, we can not only capture the distance between arbitrary cells, but also model the relationship between different modalities. Let $head[i]$ and $tail[i]$ denote the head and tail position of cell $c_i$. We use four kinds of relative distance to indicate the position information between $c_i$ and $c_j$. They can be calculated as:

\vskip -0.2in
{\small
\begin{align}
    d_{ij}^{(hh)} &=head[i]-head[j]\\
	d_{ij}^{(ht)} &=head[i]-tail[j]\\
	d_{ij}^{(th)} &=tail[i]-head[j]\\
	d_{ij}^{(tt)} &=tail[i]-tail[j],
\end{align}}

\noindent where $d_{ij}^{(ht)}$ denotes the distance between the head of $c_i$ and tail of $c_j$, and $d_{ij}^{(hh)}$, $d_{ij}^{(th)}$, $d_{ij}^{(tt)}$ have similar meanings. To obtain the position encoding $P_{pos}$ from distance value, we adopt sine and cosine functions of different frequencies as in ~\citep{vaswani2017attention}:

\vskip -0.2in
{\small
\begin{align}
	P_{pos}^{2k} &=\sin(pos/10000^{2k/d}) \\
	P_{pos}^{2k+1} &=\cos(pos/10000^{2k/d}),
\end{align}
}

\noindent where $pos$ is $d_{ij}^{(hh)}$, $d_{ij}^{(th)}$, $d_{ij}^{(ht)}$ or $d_{ij}^{(tt)}$, and $k$ is the index of dimension of position encoding. The Transformer has the same dimension $d$ as flat lattice embeddings. Then, we concatenate the four distance position encodings and feed them into a non-linear transformation layer to get the final relative position encoding of cells:

\vskip -0.2in
{\small
\begin{equation}
	R_{ij}=\operatorname{ReLU}(W_r(P_{d_{ij}^{(hh)}}\oplus P_{d_{ij}^{(ht)}}\oplus P_{d_{ij}^{(th)}}\oplus P_{d_{ij}^{(tt)}})),
\end{equation}
}

\noindent where $W_r\in\mathbb R^{d\times4d}$ is a learnable parameter, and $\oplus$ denotes the concatenation operation.

As mentioned in ~\citep{shaw2018self}, we think that commutativity of the vector inner dot will cause the loss of directionality in Transformer. Therefore, we further use a variant of self-attention ~\citep{dai2019transformer} to leverage the relative position encoding of different cells, and the attention score between the query $q_{i}$ and key vector $k_{j}$ of two semantic units can be calculated as following:

\vskip -0.2in
{\small
\begin{align}
	A_{i,j}=&E_{c_i}^T W_q^T W_{k,E} E_{c_j} +E_{c_i}^T W_q^T W_{k,R} R_{ij}\notag\\&+u^T W_{k,E} E_{c_j}+v^T W_{k,R} R_{ij},
\end{align}
}
\vskip -0.1in
\noindent where $E_{c_i}$ represents the flat lattice embedding of $i$-th cell from $x_i^c$ and $v_i^c$, or the output of last Transformer layer. We collect the attention values and embeddings of all indexes, and denote them as $A^{*}$ and $E_c^{*}$. Then, we perform self-attention over the sequence by $h=8$ parallel heads individually and $d_{head}=d/h$ is the dimension of each head. We concatenate the results and transform them into the original dimension by a linear projection. The output of the Transformer is calculated as following:

\vskip -0.1in
{\small
\begin{equation}
	Att_i=softmax(\frac{A^{*}}{\sqrt{d_{head}}})[E_c^{*}W_v]^T
\end{equation}
}

{\small
\begin{equation}
   MH\text{-}ATT=W_t[Att_1;...;Att_z],
\end{equation}
}

\noindent where $W_q$, $W_{k,R}$, $W_{k,E}$, $W_v\in\mathbb R^{d\times d_{head}}$ and $u,v\in\mathbb R^{d_{head}}$ are learnable parameters of each head, and they keep individual from different heads. $W_t\in\mathbb R^{d\times d}$ denotes the weight matrices for the multi-head attention. For simplicity, we omit the subsequent layers, which are the same as vanilla Transformer.

\paragraph{CRF Decoding.}To increase model capacity and interaction frequency, we stack $l$ Transformer layers to form a cascaded architecture. Finally, we only take the word presentation of the encoding output as $H_W\in\mathbb R^{n\times d}$, which is sent to the decoding layer for sequence labelling. Considering the dependency between successive labels, we model  $H_W$ jointly using a standard CRF layer. Let $Y^{'}$ denotes the set of all possible label sequences for input sentence $X$, the probability of the label sequence $Y$ can be calculated as:

\vskip -0.2in
{\small
\begin{equation}
	p(Y|S,O)=\frac{\prod_{n=1}^{N}\psi_n(y_{n-1},y_n,H_W)}{\sum_{y^{'}\in Y^{'}} \prod_{n=1}^{N}\psi_n(y_{n-1}^{'},y_n^{'},H_W)},
\end{equation}
}

\noindent where $\psi_n(y_{n-1},y_n,H_W)=\exp(W_{crf}H_W+b_{crf})$ is the scoring function, and $W_{crf}$ and $b_{crf}$ are the weight vector and bias. 

\subsection{Training with Entity Boundary Detection}

Since the pre-trained ResNet is intended for the image classification task, its high-level representation may overemphasize the visual objects that are prominent in the image and misidentify them as named entities. To alleviate the bias, we propose to leverage a flat text-based Transformer for entity boundary detection based on the properties of positioning scheme in the flat lattice structure. The EBD task aims to detect the position of the head and tail of entities in the input sentence, which can eliminate the types guidance from visual objects and enhance the perception of boundary words. As illustrated on the blue-frame of Figure \ref{fig:FMIT}, the flat text-based Transformer is the same structure as FMIT, but only takes the words representation as input.

\begin{table}[t]
	\caption{The statistics summary of two Twitter datasets.\label{tab:1}}
	\small
         \vskip -1em
	\centering
	\resizebox{0.48\textwidth}{15mm}{
	\begin{tabular}{@{\extracolsep{\fill}}c|ccc|ccc@{}}
		\hline
		\multirow{2}{*}{Entity Type} & \multicolumn{3}{c|}{Twitter-2015} & \multicolumn{3}{c}{Twitter-2017} \\
		&Train & Dev & Test & Train & Dev & Test \\
		\hline
		Person & 2217 & 552 & 1816 & 2943& 626 & 621 \\
		Location & 2091 & 522 & 1697 & 731 & 173 & 178 \\
		Organization & 928 & 247 & 839 & 1674 & 375 & 395 \\
		Miscellaneous &940 & 225 & 726 & 701 & 150 & 157 \\
		Total & 6176 & 1546 & 5078 & 6049 & 1324 & 1351 \\
		Num of Tweets & 4000 & 1000&3257&3373&723&723\\
		\hline
	\end{tabular}}
	\vskip -0.1in
\end{table}

We remove the type information and decompose the subsequence $Z=(z_1,z_2,...,z_n)$ from the labelling sequence $Y$, where $z_i\in\{B,E,O\}$ indicates whether the $i$-th position is the head, tail or neither of an entity. We employ the flat text-based Transformer to obtain its specific hidden representations as $T_W\in\mathbb R^{n\times d}$, followed by feeding it to another CRF layer to predict the probability of the label sequence $Z$ given $S$ as Eqn. (13):

\vskip -0.2in
{\small
\begin{equation}
	p(Z|S)=\frac{\prod_{n=1}^{N}\psi_n(z_{n-1},z_n,T_W)}{\sum_{z^{'}\in Z^{'}} \prod_{n=1}^{N}\psi_n(z_{n-1}^{'},z_n^{'},T_W)},
\end{equation}
}

In training phase, we linearly combine the loss function of the main MNER task and auxiliary EBD task, making the final training objective function by minimizing negative log-likelihood estimation as follows:

\vskip -0.2in
{\small
\begin{equation}
	\mathcal{L}=-\sum(\log p(Y|S,O)+\lambda\log p(Z|S)),
\end{equation}
}

\section{Experiments}

We conduct experiments on two MNER datasets, comparing our Flat Multi-modal Interaction Transformer (FMIT) approach with a number of uni-modal and multi-modal approaches.

\begin{table*}[t]
	\caption{Performance of different competitive text-based and multi-modal approaches on two Twitter datasets. \label{tab:2}}
	\small
         \vskip -1em
	\renewcommand{\arraystretch}{1.1}
	\setlength\tabcolsep{2.5pt}
	\centering
	\resizebox{1\textwidth}{32mm}{
		\begin{tabular}{@{\extracolsep{\fill}}c|c|cccc|ccc|cccc|ccc@{}}
			\hline
			\multirow{3}{*}{Modality} &\multirow{3}{*}{Approaches}& \multicolumn{7}{c|}{Twitter-2015} & \multicolumn{7}{c}{Twitter-2017} \\
			\cline{3-16}
			&&\multicolumn{4}{c|}{Single Type (F1)} & \multicolumn{3}{c|}{Overall}
			& \multicolumn{4}{c|}{Single Type (F1)} & \multicolumn{3}{c}{Overall}\\
            &&PER&LOC&ORG&MISC&P&R&F1&PER&LOC&ORG&MISC&P&R&F1\\
			\hline
			\multirow{4}{*}{Text}&CNN-BiLSTM-CRF&80.86&75.39&47.77&32.61&66.24&68.09&67.15&87.99&77.44&74.02&60.82&80.00&78.76&79.37\\
			&HBiLSTM-CRF&82.34&76.83&51.59&32.52&70.32&68.05&69.17&87.91&78.57&76.67&59.32&82.69&78.16&80.37\\
			&BERT&84.72	&79.91	&58.26	&38.81	&68.30	&74.61	&71.32	&90.88	&84.00	&79.25	&61.63	&82.19	&83.72	&82.95\\
			&BERT-CRF&84.74	&80.51	&60.27	&37.29	&69.22	&74.59	&71.81	&90.25	&83.05	&81.13	&62.21	&83.32	&83.57	&83.44\\
			\hline
			\multirow{8}{*}{\makecell{Text +\\ Image}}&VG-ATT&82.66	&77.21	&55.06	&35.25	&73.96	&67.90	&70.80	&89.34	&78.53	&79.12	&62.21	&83.41	&80.38	&81.87\\
			&Ada-Co-ATT&81.98	&78.95	&53.07	&34.02	&72.75	&68.74	&70.69	&89.63	&77.46	&79.24	&62.77	&84.16	&80.24	&82.15\\
			&UMT&85.24	&81.58	&63.03	&39.45	&71.67	&75.23	&73.41	&91.56	&84.73	&82.24	&70.10	&85.28	&85.34	&85.31\\
			&UMGF&84.26	&83.17	&62.45	&42.42	&74.49	&75.21	&74.85	&91.92	&85.22	&83.13	&69.83	&86.54	&84.50	&85.51\\
			\cline{2-16}
			&FMIT ($l=1$)&84.83	&83.19	&62.64	&41.13	&74.18	&75.03	&74.60	&91.75	&85.06	&82.38	&69.84	&85.55	&85.29	&85.42\\
			&FMIT ($l=3$)&\textbf{86.77}	&83.93	&\textbf{64.88}	&42.97	&75.11	&\textbf{77.43}	&\textbf{76.25}	&\textbf{93.14}	&\textbf{86.52}	&83.93	&70.90	&\textbf{87.51}	&86.08	&\textbf{86.79}\\
			&FMIT ($l=6$)&86.45	&\textbf{84.19}	&64.35	&\textbf{43.68}	&\textbf{76.28}	&75.67	&75.97	&93.04	&85.94	&\textbf{84.56}	&\textbf{71.20}	&86.80	&\textbf{86.26}	&86.53\\
			&FMIT ($l=12$)&85.79	&83.91	&62.87	&41.55	&74.92	&75.63	&75.27	&92.61	&86.03	&83.34	&70.78	&86.32	&85.50	&85.91\\
			\hline
	\end{tabular}}
	\vskip -0.1in
\end{table*}

\begin{table}[t]
	\caption{The relative inference-time speed and parameters of different models in the information interaction phrase.\label{tab:3}}
	\small
         \vskip -1em
 	\renewcommand{\arraystretch}{1.1}
	\centering
	\resizebox{0.35\textwidth}{14mm}{
	\begin{tabular}{@{\extracolsep{\fill}}c|c|c@{}}
		\hline
		Approaches & Speedup & Parameters  \\
		\hline
		UMT & 1.93$\times$ & 403M  \\
		UMGF & 1$\times$ & 231M  \\
		FMIT ($l=1$) & \textbf{27.73$\times$} &\textbf{12M}  \\
		FMIT ($l=3$) & 10.17$\times$ & 35M \\
		FMIT ($l=6$) & 6.12$\times$ & 69M\\
		\hline
	\end{tabular}}
	\vskip -0.1in
\end{table}

\subsection{Datasets}

We take two publicly benchmark Twitter datasets (Twitter-2015 and Twitter-2017) for MNER, which are provided by ~\citep{zhang2018adaptive} and ~\citep{lu2018visual}, respectively. Each sample consists of a \{Sentence, Image\} pair. Since some samples lack image modality, we replace the missing images with a uniform empty image. Table\ref{tab:1} shows the number of entities for each type and the size of train/dev/test data split.

\subsection{Baselines}

For a comprehensive comparison, we mainly compare two groups of baselines with our approach.

The first group is the representative text-based approaches for NER: (1) \textbf{CNN-BiLSTM-CRF}~\citep{ma2016end} and \textbf{HBiLSTM-CRF}~\citep{lample2016neural}, leverage both character-level information and BiLSTM-based word-level information. (2) \textbf{BERT}~\citep{devlin2019bert} and \textbf{BERT-CRF}, a pre-trained multi-layer bidirectional Transformer encoder.

The second group is several competitive multi-modal approaches for MNER: (3) \textbf{VG-ATT}~\citep{lu2018visual}, based on \textbf{HBiLSTM-CRF} with the visual context, utilizes a visual attention model and a gate mechanism to mine implicit the word-aware visual information. (4) \textbf{Ada-Co-ATT}~\citep{zhang2018adaptive}, a multi-modal approach based on \textbf{CNN-BiLSTM-CRF}, designs an adaptive co-attention network to fuse word-guided visual representations and image-guided textual representations by a filtration gate. (5) \textbf{UMT} ~\citep{yu2020improving} empowers Transformer with a multi-modal interaction module to capture the inter-modality dynamics and incorporates the auxiliary entity span detection module. (6) \textbf{UMGF} ~\citep{zhang2021multi}, the existing state-of-the-art approach for MNER, uses a unified multi-modal graph to capture the semantic relationships between the words and visual objects and stack multiple fusion layers to perform semantic interactions to learn node representations.

\subsection{Experiment Results}

We mainly adopt standard Precision (P), Recall (R) and F1-score (F1) to evaluate the overall performance on two Twitter MNER datasets and report the metric F1 for each single type. To demonstrate the effectiveness and robustness of \textbf{FMIT}, we conduct extensive experiments from self-domain and cross-domain scenarios.

\paragraph{Self-domain Scenario.} Table \ref{tab:2} shows the performance comparison of our FMIT approach with different competitive text-based and multi-modal approaches in a self-domain scenario for MNER.


\begin{table*}[t]
	\caption{Performance comparison of \textbf{FMIT} and two existing state-of-the-art multi-modal approaches in cross-domain scenarios for generalization analysis. \label{tab:4}}
	\small
	\centering
         \vskip -1em
	\setlength\tabcolsep{5pt}
	\renewcommand{\arraystretch}{1.1}
	\resizebox{1\textwidth}{14mm}{
		\begin{tabular}{@{\extracolsep{\fill}}c|cccc|ccc|cccc|ccc@{}}
			\hline
			\multirow{3}{*}{Approaches}& \multicolumn{7}{c|}{Twitter-2017$\rightarrow$Twitter-2015} & \multicolumn{7}{c}{Twitter-2015$\rightarrow$Twitter-2017} \\
			\cline{2-15}
			&\multicolumn{4}{c|}{Single Type (F1)} & \multicolumn{3}{c|}{Overall}
			& \multicolumn{4}{c|}{Single Type (F1)} & \multicolumn{3}{c}{Overall}\\
			&PER&LOC&ORG&MISC&P&R&F1&PER&LOC&ORG&MISC&P&R&F1\\
			\hline
			UMT &80.34	&71.30	&47.97	&20.13	&64.67	&63.59	&64.13	&81.24	&67.89	&39.52	&31.87	&67.80	&55.23	&60.87\\
			UMGF &79.62	&71.94	&49.48  &20.24  &\textbf{67.00}  &62.81  &66.21  &81.83  &\textbf{72.25}  &41.20  &32.00  &69.88  &56.92  &62.74\\
			FMIT$(l=3)$ &\textbf{82.05}  &\textbf{72.33}  &\textbf{50.82}  &\textbf{21.28}  &66.72  &\textbf{69.73}  &\textbf{68.19}  &\textbf{83.51}  &71.96  &\textbf{42.93}  &\textbf{33.46}  &\textbf{70.65}  &\textbf{59.22}  &\textbf{64.43}\\
			\hline
	\end{tabular}}
\end{table*}

\begin{table*}[t]
	\caption{Ablation study of our FMIT. \label{tab:5}}
	\small
	\centering
        \vskip -1em
	\setlength\tabcolsep{5pt}
	\renewcommand{\arraystretch}{1.1}
	\resizebox{1\textwidth}{15mm}{
		\begin{tabular}{@{\extracolsep{\fill}}c|cccc|ccc|cccc|ccc@{}}
			\hline
			\multirow{3}{*}{Approaches}& \multicolumn{7}{c|}{Twitter-2015} & \multicolumn{7}{c}{Twitter-2017} \\
			\cline{2-15}
			&\multicolumn{4}{c|}{Single Type (F1)} & \multicolumn{3}{c|}{Overall}
			& \multicolumn{4}{c|}{Single Type (F1)} & \multicolumn{3}{c}{Overall}\\
			&PER&LOC&ORG&MISC&P&R&F1&PER&LOC&ORG&MISC&P&R&F1\\
			\hline
			FMIT ($l=3$)&\textbf{86.77}	&\textbf{83.93}	&\textbf{64.88}	&\textbf{42.97}	&\textbf{75.11}	&77.43	&\textbf{76.25}	&\textbf{93.14}	&\textbf{86.52}	&\textbf{83.93}	&\textbf{70.90}	&\textbf{87.51}	&86.08	&\textbf{86.79}\\
			w/o Obj&85.55	&80.73	&63.37	&38.61	&73.44	&74.25	&73.84	&92.16	&85.23	&81.57	&68.97	&85.36	&84.69	&85.02\\
			w/o Rel&84.13	&79.95	&62.64	&38.86	&72.87	&73.38	&73.12	&91.06	&85.36	&81.24	&67.83	&84.29	&84.76	&84.52\\
			w/o EBD&86.21	&83.26	&64.05	&42.21	&73.38	&\textbf{77.95}	&75.60	&92.73	&86.14	&82.55	&69.47	&85.55	&\textbf{86.67}	&86.11\\
			\hline
	\end{tabular}}
	\vskip -0.1in
\end{table*}

(1) Compared with the text-based approaches, the multi-modal approaches can generally achieve better performance than their corresponding uni-modal baselines, which demonstrates that incorporating the visual information is motivating for NER in social media. For example, in the overall F1 on both datasets, \textbf{VG-ATT} outperforms \textbf{HBiLSTM-CRF} by 1.63$\%$ and 1.50$\%$, respectively. Moreover, recent multi-modal approaches \textbf{UMT} and \textbf{UMGF} show significant performance improvements when replacing the sentence encoder with BERT and using the Transformer framework to interact textual and visual information. It further demonstrates that the self-attention in Transformer is more beneficial for feature fusion and filtering.

(2) Compared with \textbf{UMT} and \textbf{UMGF}, which utilize Transformer to model intra-modal and inter-modal information interactions and dynamically control the contribution of visual features through a gating mechanism, Our \textbf{FMIT} makes radical promotion in model structure and representations of different modalities. In the overall F1 on both datasets, our best model achieves state-of-the-art performance by obtaining 76.25$\%$ and 86.79$\%$ results, outperforming \textbf{UMT} by 2.84$\%$ and 1.48$\%$, and outperforming \textbf{UMGF} by 1.40$\%$ and 1.28$\%$, respectively.

(3) We compare the impact of the number of Transformer layers. It can be observed that when $l=1$, \textbf{FMIT} achieves performance comparable to \textbf{UMT} and \textbf{UMGF}, both of which use the 12-layers Transformer framework. When $l=3$ or $l=6$, we reach state-of-the-art F1-score in all single types and overall F1-score, precision and recall metrics on both datasets. As shown in Table \ref{tab:3}, our approach can achieve better performance with fewer parameters and higher efficiency. With only 12M parameters, the 1-layer \textbf{FMIT} is 14.37 times and 27.73 times faster than \textbf{UMT} and \textbf{UMGF} in information interaction phases, respectively. It demonstrates that flat lattice structure can more fully and directly establish interactions in both intra-modal and inter-modal simultaneously, making it possible to incorporate important visual information into entity recognition with only a few Transformer layers. Meanwhile, we speculate that the reason for the performance declines at 12-layer \textbf{FMIT} is the redundancy of unnecessary information.

\paragraph{Cross-domain Scenario.} Due to the obvious differences in type distribution and data characteristics between the two Twitter datasets, we compare our \textbf{FMIT} approach and two existing state-of-the-art multi-modal approaches in cross-domain scenarios for generalization analysis. Twitter-2017$\rightarrow$Twitter-2015 indicates that the model trained on Twitter-2017 is used to test Twitter-2015, and Twitter-2015$\rightarrow$Twitter-2017 has similar meaning. As shown in Table \ref{tab:4}, our approach outperforms \textbf{UMT} and \textbf{UMGF} by a large margin in most metrics. The potential reason for the excellent generalization may be that with the tight information coupling structure enables \textbf{FMIT} to learn the underlying features better.

\subsection{Ablation Study}

To investigate the influence of different factors of our proposed approach, we perform comparison between the 3-layer \textbf{FMIT} and its ablation approaches, concerning the entity boundary detection task and several critical components of the model. The results are reported in Table \ref{tab:5}.

\begin{figure*}[t]
\begin{center}
\vskip -0.1in
\centerline{\includegraphics[width=0.95 \linewidth]{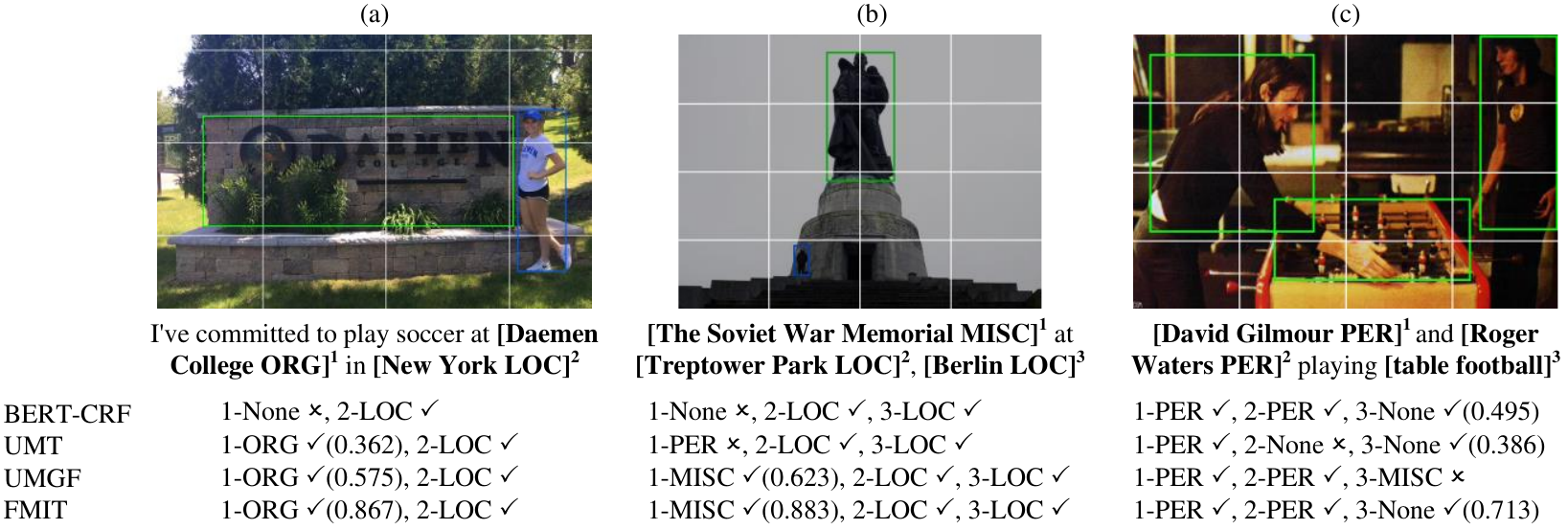}}
\vskip -0.1in
\caption{ The first row shows several representative samples together with their manually labeled entities in the test set of two Twitter datasets, and the bottom four rows show predicted entities of different approaches on these test samples. The values in parentheses represent the confidence of the predicted label.}\label{fig:3}
\end{center}
\vskip -0.3in
\end{figure*}

\paragraph{w/o Obj.} Firstly, we replace the targeted visual object guidance with $7\times7$ average-segmented visual blocks, which can be obtained by feeding the whole image to ResNet~\citep{he2016deep} and taking the output of the last convolution layer. This approach completely ignores the correspondence of fine-grained units between different modalities, bringing in significant performance degradation.

\paragraph{w/o Rel.} Secondly, we remove the relative position encoding for flat lattice structure and the positioning scheme of each cell. In this case, we only use self-attention in vanilla Transformer to conduct intra-modal and inter-modal fusions. We find that the overall F1 on both datasets decreases substantially by 3.13$\%$ and 2.27$\%$, respectively, which indicates a critical role for coupling interactions between different modalities through position strategy.

\paragraph{w/o EBD.} Discarding the entity boundary detection task and only retaining the main MNER task will lead to significant performance degradation in overall precision, while a slight increase in overall recall. The result is consistent with our hypothesis that the guidance of visual objects drives the corresponding words to be misjudged as entities, while the EBD auxiliary task can balance the play of visual objects and text itself.

\subsection{Further Analysis}

\begin{figure}[t]
\begin{center}
\vskip -0.1in
\centerline{\includegraphics[width=0.95 \linewidth]{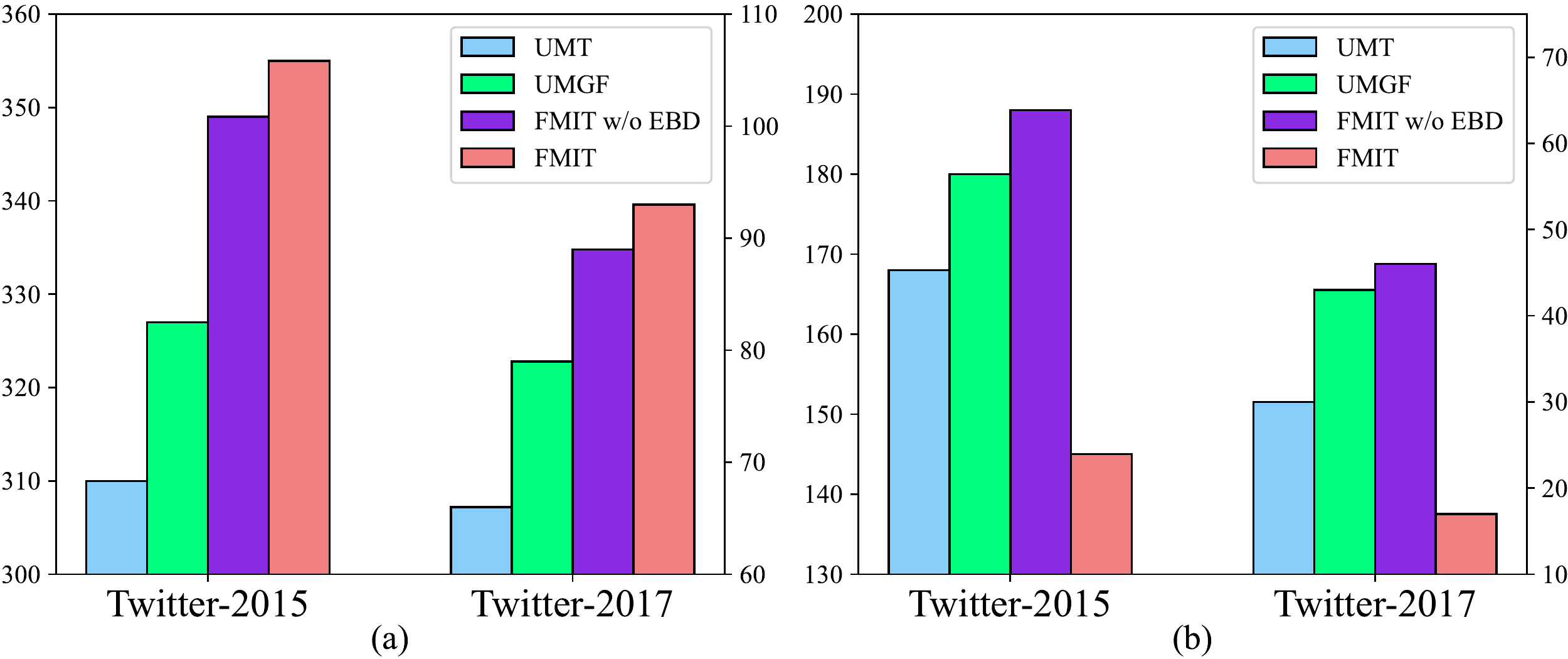}}
\vskip -0.1in
\caption{Predicted results statistics: (a) The number of entities (shown in y-axis) that are incorrectly predicted by \textbf{BERT-CRF}, but get corrected by each multi-modal approach; (b) The number of entities (shown in y-axis) that are correctly predicted by \textbf{BERT-CRF}, but wrongly predicted by each multi-modal approach.}\label{fig:4}
\end{center}
\vskip -0.3in
\end{figure}

\paragraph{Case Study.} To better understand the effective of our approach in incorporating visual information into the MNER task, we select a representative set of test samples to compare the prediction results of the 3-layer \textbf{FMIT} and other approaches.

First, from Figure \ref{fig:3}(a), we can observe that the \textbf{BERT-CRF} fails to identify \textit{Daemen College} due to the lack of guidance from visual context such as the plaque, while all the multi-modal approaches can accurately determine the entities by referring to specific visual regions.

Second, we can see from Figure \ref{fig:3}(b) that \textbf{UMT} gives a wrong identification of the entity \textit{The Soviet War Memorial}, probably because the segmented visual feature is fragmented, bringing in interference to type classification. On the contrary, \textbf{UMGF} and \textbf{FMIT} can accurately classify the entities into corresponding types with the guidance of targeted visual objects.

Third, as shown in Figure \ref{fig:3}(c), \textbf{UMGF} erroneously identifies \textit{table football} as an entity of MISC, which indicates that over-reliance on visual information will lead to emphasis bias. Therefore, \textbf{FMIT} corrects this bias by balancing the importance of text and vision with EBD task.

Finally, we find that compared with other approaches, \textbf{FMIT} can obtain higher confidence in predicted results. For example, for entity \textit{Daemen College} in Figure \ref{fig:3}(a), \textbf{FMIT} achieves a label confident of 0.867, substantially outperforming \textbf{UMGF}(0.575) and \textbf{UMT}(0.362). It indicates that our relative position encoding strategy and flat lattice structure can extract important information and couple different modalities more directly.

\paragraph{Statistic Study.} To better appreciate the importance of the EBD auxiliary task, we count the number of entities that are correctly/wrongly predicted by \textbf{BERT-CRF}, but wrongly/correctly predicted by each multi-modal approach.

As shown in Figure \ref{fig:4}, compared with other multi-modal methods, our \textbf{FMIT} can correctly identify more entities due to the powerful image-ware word representations. Moreover, it is clear that \textbf{FMIT} introduces fewer wrong entities with the help of EBD auxiliary task. It demonstrates that the well-designed EBD auxiliary task can greatly eliminate the visual bias brought by visual context and perform more efficiently than the span detection module proposed in \textbf{UMT}.

\section{Conclusion}

In this paper, we propose a novel Flat Multi-modal Interaction Transformer for MNER, which exploits flat lattice structure and relative position encoding to directly interact fine-grained semantic units between different modalities. Moreover, we put forward entity boundary detection as an auxiliary task to alleviate visual bias. We conduct extensive experiments on two MNER datasets, and experimental results demonstrate that our approach outperforms other text-based and multi-modal approaches.

\bibliography{anthology,custom}
\bibliographystyle{acl_natbib}

\appendix

\section{Implementation Details}
\label{sec:Implementation Details}

 For each uni-modal and multi-modal approach, the maximum length of the sequence input and batch size are respectively set to 128 and 16. For our FMIT approach, we utilize the pre-trained cased $BERT_{base}$ model with dimension of 768 to initial word representations $H_x$, and employ a pre-trained 152-layer ResNet with dimension of 2048 to initial the visual representations $H_v$. The parameters of both pre-trained models keep fine-tuned during training. After dimension aligned, the dimension $d$ of both modalities are transformed into 512. The dropout rate and tradeoff rate $\lambda$ are respectively set to 0.2 and 0.25. To train our model, we use Adam optimizer with a learning rate of 5e-5 for pre-trained models and 2e-4 for other parameters. 

\end{document}